\documentclass[10pt,letterpaper]{article}

\usepackage{cogsci}

\cogscifinalcopy % Uncomment this line for the final submission 

\usepackage{pslatex}
\usepackage[natbibapa]{apacite}
\usepackage{float} % Roger Levy added this and changed figure/table
\usepackage{graphicx}
\usepackage{algorithm}
\usepackage{algorithmic}
\usepackage{array}
\usepackage{pgfplots}
\usepackage{pgfplots, pgfplotstable}
\usepackage{amsmath}
\usepackage{verbatim}
\title{Givenness Hierarchy Theoretic Cognitive Status Filtering}
 
 \author{ \textbf{Poulomi Pal (poulomipal@mines.edu), Lixiao Zhu, Andrea Golden-Lasher,}\\ \textbf{Akshay Swaminathan, and Tom Williams} \\
   MIRRORLab, Colorado School of Mines\\
   1600 Illinois Street, Golden, CO 80401 USA}
\begin{document}

\maketitle

\begin{abstract}
For language-capable interactive robots to be effectively introduced into human society, they must be able to naturally and efficiently communicate about the objects, locations, and people found in human environments. An important aspect of natural language communication is the use of pronouns. According to the linguistic theory of the \textit{Givenness Hierarchy} (GH), humans use pronouns due to implicit assumptions about the \textit{cognitive statuses} their referents have in the minds of their conversational partners. In previous work, Williams et al. %\citep{williams2018reference, williams2018towards, williams2016situated} 
presented the first computational implementation of the full GH for the purpose of robot language understanding, leveraging a set of rules informed by the GH literature.
However, that approach was designed specifically for language understanding, oriented around GH-inspired memory structures used to assess what entities are candidate referents given a particular cognitive status. In contrast, language generation requires a model in which cognitive status can be assessed for a given entity. We present and compare two such models of cognitive status: 
%
%In contrast, no model of Givenness Hierarchy threoretic robotic language generation has been developed yet. Specifically, no computational model of cognitive status yet exists. Thus, we present \textit{two} models of cognitive status, both structured so as to be optimized for natural language generation rather than natural language understanding; 
a rule-based \textit{Finite State Machine} model directly informed by the GH literature and a \textit{Cognitive Status Filter} designed to more flexibly handle uncertainty. The models are demonstrated and evaluated using a silver-standard English subset of the OFAI Multimodal Task Description Corpus.

\textbf{Keywords:} 
Cognitive Status modeling, Natural language generation, Human-robot interaction
\end{abstract}

\section{Introduction}

As human-robot interaction becomes increasingly common, robots need to be able to talk about the objects, locations, and people in their environments in the same way humans do, to facilitate concise, easy, and unambiguous communication. 
%The use of \textit{pronouns} is an important aspect of natural language communication between humans. 
To reap these benefits, just like humans, robots must be able to understand and use pronouns like \textit{it}, \textit{this}, and \textit{that}. The linguistic theory of the \textit{Givenness Hierarchy} (GH) \citep{gundel1993cognitive} suggests that humans tend to use pronouns rather than longer referring expressions due to implicit assumptions about the \textit{cognitive status} the referent has in the mind of their interlocutor. That is, the use of different referring forms is viewed as justified based on whether the referent is \textit{In Focus}, \textit{Activated}, \textit{Familiar}, and so forth, within the current conversation. Thus, for robots to understand and generate human-like natural language they must be able to model this notion of cognitive status.

%The knowledge of cognitive status of an entity is important for both understanding and generation of natural language and thus robots need models of cognitive status that facilitate both understanding and generation. 
Previously, \citet{williams2018reference} (see also \citep{williams2018towards, williams2016situated}) presented the first full computational implementation of the GH for the purpose of robotic natural language understanding, using a set of hand-crafted rules informed by the GH literature.
However, that approach was designed specifically for robotic natural language understanding, oriented around GH-inspired memory structures used to assess what entities are candidate referents given a particular cognitive status. In contrast, natural language generation requires a model in which cognitive status can be assessed for a given entity.
%Working within this theory, we argue that if robots have to learn and decide which pronominal forms to use for effective communication with humans, they must be able to model the \textit{cognitive status} of their target referents within their current conversation.

%Their approach was designed specifically for natural language understanding, oriented around GH-inspired memory structures containing the entities the robot believes its interlocutor to believe that it could plausibly believe to have each cognitive status (e.g., the set of entities the robot believes interlocutors would believe that the robot believes to be in focus -- a second-level theory-of-mind model). While \citet{williams2018reference} demonstrated the successful use of this model for effective and efficient anaphora \textit{understanding}, we argue that it cannot be used as-is for anaphora \textit{generation}.

Such a model of cognitive status could either be developed as a rule-based model (not dissimilar from the rule-based approach to GH-theoretic language understanding taken by \citet{williams2018reference}), or could instead be developed as a statistical model which would attempt to learn to predict an entity's cognitive status from data. While in practice both rule-based and data-driven empirical models are useful \citep{bangalore2005probabilistic}, 
% rule-based models may be brittle, 
% %
% %and require continuous supervision, 
%empirical 
data-driven models 
%are robust and uses learning methods that do not require following specific rules to be 
may be better able to handle unseen, uncertain situations \citep{bangalore2003balancing, bangalore2005probabilistic}.

In this paper, we thus propose (and compare to a rule-based Finite State Machine (FSM) model) the \textit{Cognitive Status Filter (CSF)}: a data-driven probabilistic model of cognitive status, structured to be optimized for natural language generation rather than natural language understanding, trained and evaluated using a silver-standard\footnote{This subset constitutes English transliteration of originally German dialogues.} English subset of the OFAI Multimodal Task Description Corpus \citep{schreitter2016ofai}. 
Specifically, the CSF seeks to predict the cognitive status for a given entity based on whether and how it has been referenced in natural language.
%The CSF is a Bayes Filter which recursively estimates for a given entity the probability distribution over cognitive status for that entity, based on the previous estimated distribution over cognitive statuses and the linguistic features observed for that entity in the dialogue at the previous timestep.

%%%%%EDITED TO HERE%%%%

% Since the GH provides a connection between the GH's hierarchically nested tiers of cognitive statuses and the possible referring forms associated with each tier, we also wanted to explore whether taking a data-driven probabilistic approach would be able to capture the same connection with more nuance. 

%

% Our goal is to learn the statistical relationships between when people use concise forms like ``it", ``this" or ``that" and why they are using them, to allow us to track what listeners likely think is ``in focus" or ``activated" within the conversation. Thus, designing this probabilistic model of cognitive status will provide us with the opportunity to develop a computational model of reference generation that is based on the GH and can also handle uncertainty when applied to real-life scenarios.

The remainder of this paper is organized as follows. After discussing related work on cognitive status and referring expressions, we formally define the concept of a Cognitive Status Filter. We then present the results of a crowdsourced human-subject experiment to gather the data necessary to train and evaluate this model, and compare the CSF model's performance to that of a rule-based Finite State Machine model. Finally, we discuss our results and conclude with possible directions for future work.

\section{Related Work}

The Givenness Hierarchy, originally presented by \citet{gundel1993cognitive}, consists of a nested hierarchy of six tiers of cognitive status: \textit{\{in focus $\subseteq$ activated $\subseteq$  familiar $\subseteq$  uniquely identifiable $\subseteq$  referential $\subseteq$  type identifiable\}}, each of which is associated with a set of referring (or pronominal) forms that can be used when referring to an entity with that status \citep{gundel2006coding,hedberg2013applying}. The hierarchical nesting here means that an entity with one status can also be said to have all other statuses lower in the hierarchy. If a target referent is \textit{in focus}, for example, it can also be inferred to be activated, familiar, and so forth. Accordingly, a speaker's selection of a pronominal form depends on their assumptions as to the cognitive status of their target referent in the mind of their conversational partner. For example, if a speaker uses ``it'' to refer to an object, the listener can %(and implicitly does) 
infer that the object being referenced must be one that is already  \textit{in focus}, whereas if a speaker uses ``that $<NP>$'', the speaker can only infer that the object is at least familiar (but may in fact be activated or even in focus). 

The hierarchical structure of the GH is also important due to the way it parallels the hierarchical nesting of models of human memory, such as \citet{cowan1998attention}'s, in which the focus of attention is a subset of short-term memory (or working memory), which is in turn a subset of long-term memory. % \citep{williams2018reference}.

The GH \textit{coding protocol}, presented by \citet{gundel2006coding}, provides guidelines as to what features of linguistic and environmental context should dictate the cognitive status of a given entity. For example, this protocol suggests that an entity that is mentioned in a topic role in the preceding clause should be considered to be in focus, and that any entity that is mentioned at all should be considered to be at least activated~\citep{gundel2006coding, hedberg2013applying}. 
%For example, if an object is the main topic of a sentence (one of the requirements of the protocol for an object to have the ``in focus" status), then the linguists say it should be regarded as \textit{in focus of attention} within the current conversation and will most likely be referred to by \textit{it} in the following sentence.

Due to the GH's popularity within the research literature, and its validation across a wide variety of languages beyond English \citep{gundel2010testing}, many researchers have sought to computationally implement it in whole or in part, especially within the context of reference resolution algorithms. \citet{kehler2000cognitive}, for example, use the GH to justify an approach in which elements of an interface that are highlighted are considered to be "in focus", and referring expressions that use pronominal forms are automatically resolved to those highlighted referents.

% Humans interacting with this system generated referring expressions depending on what was highlighted (``in focus") or not (``activated") within the visual context of the system interface. Thus, the system implicitly conveyed the likely cognitive status of its objects and, depending on that, humans chose the referring forms like ``it", ``this", ``that", etc. to indicate their target referents. Thus, using a simple set of rules along with a visual context, the system was able to perform reference resolution with high accuracy.

%However, this simple heuristic approach was only effective for single referring expressions coupled with precise gestures. 
Building on this work, \citet{chai2004probabilistic} proposed a probabilistic graph-matching algorithm for resolving referring expressions that are complex (involving multiple target referents) and ambiguous (involving gestures that could indicate multiple candidate referents) in multimodal user interfaces. Because this algorithm had high computational complexity, \citet{chai2006cognitive} demonstrated how the algorithm's performance could be improved using a greedy algorithm based on the theories of Conversational Implicature \citep{grice1975logic, dale1995computational} and the GH. Chai et al. combine these theories to create a reduced hierarchy: \textit{\mbox{Gesture $\subseteq$ Focus $\subseteq$ Visible $\subseteq$ Others}}, where Focus combines the ``in focus" and ``activated" tiers of the GH, and Visible combines its ``familiar" and ``uniquely identifiable" tiers. When a referring expression is processed, the %statistical 
relationship between %its %the used %use of certain 
referring form and status %within this reduced hierarchy 
is then used to help resolve that referring expression.

% It was demonstrated that the greedy algorithm using cognitive principles outperformed the greedy algorithm not using them by more than 15\%, indicating that modeling the cognitive status along with its associated referring forms can likely improve the robustness of interpretation of referring expressions in multimodal scenarios \cite{chai2006cognitive}.

Finally, while the approaches above focused on modeling of reduced versions of the GH, \citet{williams2016situated, williams2018reference} instead presented an implementation of the full GH, through a set of rules that associated different referring forms with different sequences of actions involving all six tiers of the GH. 
%They demonstrated how this approach better enabled natural language understanding in uncertain and open-world scenarios. 
This required, in part, four data structures corresponding to the top four tiers of cognitive statuses of the GH, while the last two tiers were instead associated with new ``mnemonic actions" such as creating new mental representations~\citep{williams2018reference}. 
%
% The algorithm exploits these data structures in two ways during reference resolution. First, selecting a particular data structure to search for and second, selecting the appropriate target referent among the other entities present in the same data structure. While the former depends on the referring form used in a referring expression, the latter depends on the ``degree of salience and uncertainty" associated with the entities .

In all of these previous approaches, the GH is used to justify a set of data structures used to store representations for entities that could be referred to, and to justify which of these data structures should be considered (and how) when a given referring form is used. However, while this is sensible during natural language understanding, it may not be appropriate for the purposes of natural language generation. During generation, the speaker already knows what object they wish to refer to, and do not need to search through these sorts of data structures. Instead, when a speaker decides what referring form to use to refer to a given object, we argue that they would instead start by determining the status of that object, and only then may they look through the data structure associated with that status, in order to determine what distractors must be ruled out. Critically, this requires %as a key capability 
the ability to quickly determine the cognitive status of a given entity. Accordingly, in the next section we propose an approach to this problem, which we term as \textit{cognitive status modeling}.

\section{Problem Formulation}

We formulate cognitive status modeling as a Bayesian filtering problem. Let a dialogue $D$ consist of a set of utterances $U_0,...,U_n$. For object $o$, let $S_{o}^{t} \in \{I,A,F\}$ denote the cognitive status of $o$ at a particular timestep $t$ after utterance $U_t$ (either In Focus, Activated or Familiar), and let $L_{o}^{t} \in \{N,M,T\}$ denote the linguistic status of $o$ in utterance $U_{t}$ (e.g., either not mentioned in the utterance, mentioned in the utterance in a non-topic role, or mentioned in the utterance in a topic role). Using this formalism, our goal is to recursively estimate, for a given object, the probability distribution over cognitive statuses for object $o$ at time $t$:

\begin{equation}
    p(S_o^t) = p(S_o^{t-1})p(L_o^{t})p(S_o^t \mid S_o^{t-1}, L_o^{t}) \label{eq:1}
\end{equation}
 
We define a Bayesian filter of this form as a \mbox{\textit{Cognitive Status Filter}} (CSF) for a given object $o$. Given a set of known objects, $O = \{o_1,...,o_n\}$, our goal is then to estimate this distribution for each $o \in O$ at each time step. To do so, we use a Cognitive Status Modeling Engine $C$, consisting of a set of CSFs $\{c_0,...,c_1\}$, one for each object believed to be of a status familiar or higher within the conversation. 
%Within this framework, we note how each GH-theoretic cognitive status is or is not tracked. 
Here, we make the simplifying assumption that the same set of objects are known to both the robot and its conversational partner, meaning that the set of all objects with status \textit{Uniquely Identifiable} or higher is simply the set of objects $O$. We assume that it is straightforward to determine whether one of these objects is or is not \textit{Familiar} based on whether or not it has appeared in the current conversation. This allows us to model whether or not an object is of status Familiar or higher based on whether or not a CSF $c\in C$ exists for that object,
and to model \textit{which} of those statuses the object likely has, using its associated CSF.
%Finally, we estimate whether each of the higher statuses (Activated, In Focus) hold for the object using the CSF $c$. 

%Mathematically, this model represents the probability of a particular cognitive status of an object at the current time stamp given the cognitive status and the linguistic (appearance) feature observed for the object on the previous time stamp, i.e, $p(S_o^t\mid S_o^{t-1}, A_o^{t-1})$.

\section{Data Collection}
The core component of our CSF model that must be learned ahead of time is the conditional probability $p(S_o^t\mid S_o^{t-1}, L_o^{t})$. To learn this, we trained our model using a silver-standard English translation of the German OFAI Multimodal Task Description corpus \citep{schreitter2016ofai}. The corpus represents a collection of human-human and human-robot interactions where the human teacher shows and explains to a human or robot learner how to connect two separate parts of a tube and then how to mount the tube onto a box with holders, as shown in Figure~\ref{fig 1} by actually moving around the objects and performing the task while explaining it to the learner. The average length of a sentence that is used in this corpus has 8-9 words. As the name suggests, since the corpus is ``multimodal", the corpus contains both verbal and non-verbal cues such as speech, gaze, and gestures. Realistic multimodal HRI scenarios require the use of such non-verbal cues; %with our uncertainty sensitive model, 
however as our first step we begin in this work by looking only at our model's ability to handle the same kind of linguistic factors %(even a simple subset of the linguistic factors) 
that are handled by the GH, leaving the ability to model other linguistic factors for future work. 

%. In order to make the corpus more diverse the authors use different people for the role of the teacher, to capture how different people might understand and represent information differently for the same task \citep{schreitter2016ofai}. 
While the OFAI MTD corpus contains data from four task scenarios, we only use the data from one particular task scenario (Task 3). The original dataset for this task consists of 16 monologues each having approximately 4 to 5 utterances. As a first step, in this work we begin by evaluating our model on a small subset of the original dataset, consisting of 4 of these monologues, each of which is comprised of just 4 utterances, to control for monologue length. As shown in Figure \ref{fig 1}, this task context contains 8 objects, including the learner and teacher. 

Task 3 was selected because it includes a larger number of objects than the other tasks in a dyadic instruction context, and contains data from both human-human and human-robot dyads. Specifically, Task 1 involved a human teacher explaining and performing a task in front of the camera without the presence of a learner in the scenario; Task 2 involved a human teacher and a human learner jointly performing the task of moving an object; and Task 4 is a pure ``navigation task" involving both human-human and human-robot dyads \citep{schreitter2016ofai}.

\subsection{Appearance Feature Annotation}
To collect linguistic status information \textit{L}, three annotators independently annotated the OFAI Multimodal Task Description Corpus \citep{schreitter2016ofai} according to the following annotation procedure. Each annotator was provided a printed copy of all 16 monologues to annotate. For each sentence in each monologue, the annotator was instructed to underline any piece of the text that could refer to some object in the scene. For each of these underlined pieces of text, the annotator was instructed to indicate the correspondence between the underlined sentence fragment and the object in the scene it referred to. Finally, the annotator was required to circle the fragment-object mapping they believed to be the topic of the sentence. There were a few cases in which annotators circled multiple objects as the topic of the sentence; in these cases, both objects were recorded as being equally probable topic referents\footnote{The inter-annotator agreement score as measured through Fleiss' Kappa was $\kappa_n$ = 0.37, indicating fair agreement between annotators. It will be important in future work to adapt the annotation protocol to increase rate of agreement.}.

\subsection{Cognitive Status Annotation}
Ground-truth cognitive status information was then collected through a crowdsourced human-subject experiment. 160 US participants were recruited from Amazon Mechanical Turk. Two participants answered an attention check question incorrectly and were dropped from our analysis, leaving 158 participants (71 female, 85 male, 2 N/A). Participant ages ranged from 19 to 70 years (M = 35.03, SD = 11.36). Each participant was paid \$0.25 for completing the study. 

\subsubsection{Procedure:}
% We started by creating a subset of the original dataset (as mentioned earlier) and then converted them to audios by having a trained experimenter read each utterance in each monologue, thereby creating 16 audios in total. 

\begin{figure}[h]
\includegraphics[width=8.5cm]{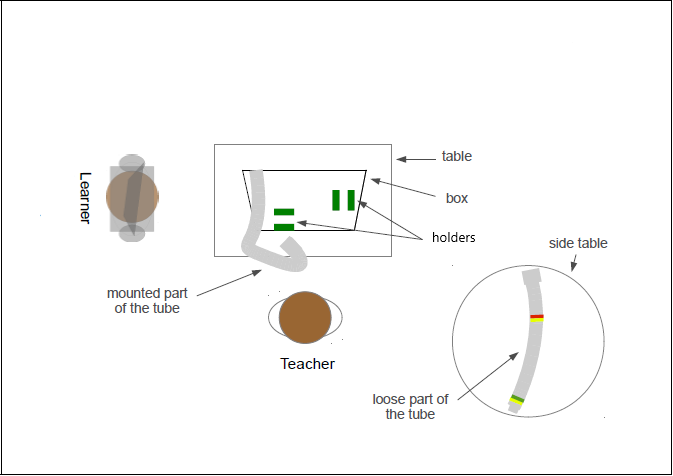}
\caption{Scene (labeled)}
\label{fig 1}
\end{figure}

At the beginning of the experiment, each participant is shown the scene depicted in Figure \ref{fig 1}, and is instructed to remember the objects and their labelings in order to performing their upcoming task. Participants were then shown the same scene without labels while listening to a portion of one of the experiment's four monologues, as read by the experimenters. Specifically, participants were randomly assigned to hear a random prefix of a randomly selected monologue (i.e., either only the first utterance of that monologue, the first two, the first three, or all four).

% one of the four monologues, for that experiment, were randomly assigned to hear either only the first utterance, the first two utterances, the first three utterances, or all four utterances from that monologue.

%%%%%%%%%%%%%%%%%%%%%%%%%%%%

% that requires them to click on one or all the objects with respect to two questions depending on what they heard. Each participant listened to a recording of an experimenter reading a subset of one of the four monologues, with participants evenly distributed between monologues, and within each monologue, participants evenly distributed between hearing only the first clause (first utterance), the first two clauses together (second utterance), the first three clauses together (third utterance), and the first four clauses together (fourth utterance).

% \begin{figure}[h]
% \includegraphics[width=7cm]{Scene_main.png}
% \caption{Scene (unlabeled)}
% \label{fig 2}
% \end{figure}

At the end of this monologue excerpt, participants were asked to answer two questions, presented in a randomized order, with the second question becoming available after the first question was answered. The two questions are as follows:
\begin{itemize}
    \item \textbf{Q1}: \textit{Click on the object in the scene that you think the speaker would most likely be referring to if the speaker would have said ``look at it'' at the end of the monologue.} 
    \item \textbf{Q2}: \textit{Click on all the objects in the scene that you think the speaker would most likely be referring to if the speaker would have said ``look at that'' at the end of the monologue.}
\end{itemize}

Two of the monologues used in our experiment are shown below. 
%After each utterance of the first monologue the question Q1 was asked whereas for the other monologue, question Q2 was asked.

\begin{description}\itemsep0em
\item [Monologue 1:]
\item [U1:] You must take the tube with your right hand.
\item [U2:] %You must take the tube with your right hand. 
And insert it in at the yellow-green connection here.
\item[U3:] %You must take the tube with your right hand. And insert it in at the yellow-green connection here. 
Put it on the tube.
\item[U4:] %You must take the tube with your right hand. And insert it in at the yellow-green connection here. Put it on the tube. 
Again, with your right hand insert it here in the holder.
\end{description}

\begin{description}
\item [Monologue 2:]\itemsep0em
\item [U1:] With the right hand stick the two tubes together.
\item [U2:] %With the right hand stick the two tubes together. 
You put that together here with the yellow-green mark.
\item[U3:] %With the right hand stick the two tubes together. You put that together here with the
%yellow-green mark. 
It is okay that it is not holding firmly.
\item[U4:] %With the right hand stick the two tubes together. You put that together here with the yellow-green mark. It is okay that it is not holding firmly. 
Now lead the one tube through here.
\end{description}

These questions allowed us to probe the user's implicit beliefs as to the cognitive status of the objects in the scene. From a GH-theoretic perspective, if a participant implicitly believed a given object to be in focus, they should click on that particular object for both \textbf{Q1} and \textbf{Q2}, whereas if they believed the object to be activated, they should click on that object for \textbf{Q2} but not for \textbf{Q1}. 
Because the context is narrowly defined and participants were given time to examine each object in the scene, we assume that all objects in the scene are familiar or higher. Thus, if a participant believed the object to be familiar or lower, they should not click on the object at all.  
%Thus, if an object is not clicked on for either question, we take this as evidence that they subconsciously believe that object to be familiar but no higher. 
After completing the task, participants completed a check question (cf. \citet{schreitter2016ofai}) requiring users to identify the scene they had viewed from among several distractors. This allowed us to ignore data from participants who did not pay sufficient attention while completing the task.

Using this coding procedure, we are thus able to determine the perceived cognitive status of each object in the scene for each participant after the completion of the monologue excerpt they were exposed to. When paired with the linguistic status annotations, this allowed us to train our CSF model, using the procedure described in the following section. %according to equation \eqref{eq:1}.

%Mathematically, the cognitive status of a particular object is obtained by the following equation:
%\begin{equation}
%   p(S^{t}) = p(S^{t-1}) p(A^{t}) p(S^t\mid S^{t-1}, A^t)
%\end{equation}
%We train our model on the entire data set to obtain the transition probability values for each possible pair of cognitive status and linguistic feature (S, A). 

\section{Training and Evaluation}

\subsection{Training}
After collecting this dataset, our CSF was trained in the following way: 
First, we initialized a 9x3 matrix whose rows correspond to the nine cognitive/linguistic status pairs an object could have at time $t-1$ ($(I_{t-1},N_{t})$, $(I_{t-1},M_{t})$, $(I_{t-1},T_{t})$, $(A_{t-1},N_{t})$, $(A_{t-1},M_{t})$, $(A_{t-1},T_{t})$, $(F_{t-1},N_{t})$, $(F_{t-1},M_{t})$, $(F_{t-1},T_{t})$), and whose columns correspond to the three cognitive statuses that object could have at time $t$ ($I_t$, $A_t$, $F_t$).

For each pair of adjacent utterances in each monologue $(U_{t-1},U_{t})$, we consider the data from all participants (for all objects) who provided data immediately following utterance $U_{t-1}$, and from all participants who provided data immediately following utterance $U_{t}$. For each resulting pair of datapoints, we identify and increment the correct cell in this matrix. For example, for the combination of a datapoint from a participant who heard some utterance and subsequently viewed that object as in focus, and a datapoint from a participant who heard the next utterance in the same monologue, containing object 1 in a non-topic role, and at that point viewed the object as being activated, we would increment the cell $((I_{t-1},N_{t}),A_t)$. Once all data has been considered, we normalize each row of this table to produce a conditional probability table.

\subsection{Evaluation}
To evaluate our CSF model, we then considered each object $o$ and  each monologue $M$, and retrained our model using all data except that which was collected for object $o$ or monologue $M$ (for example, while testing for object $o_1$ in monologue $M_1$, we retrain our model with all the data except that concerned with $M_1$ and/or $o_1$), and used this model (along with a prior distribution over cognitive statuses for that object as described below) to simulate what status would be predicted for that object at each point in that monologue. After each of these utterances, we evaluated the model's prediction by comparing it to the majority opinion from participants who \textit{had} provided data for that object at that point in that monologue. Combining these prediction results for all eight objects in all four utterances in all four monologues produced a 128-element prediction vector for the model.

Specifically, we computed these prediction vectors for each of two CSF models, each of which used a different prior distribution $p(S_o^{t-1})$ over cognitive statuses: 

\begin{description}
\item[U-Model:] an \textit{uninformed} prior in which each cognitive status was assigned a prior probability of 0.33.
\item[I-Model:] a (weakly) \textit{informed} prior, in which the three cognitive statuses were assigned prior probabilities {I= 0.05, A= 0.1, F= 0.85}. These probabilities reflect the fact that objects are a priori far more likely to be familiar than activated, and among the set of things that are currently activated it is more likely for a given object to be activated than in focus. While in theory this distribution could be learned from data, in a realistic environment it may be the case that hundreds or thousands of objects are familiar and only one is in focus, yielding an extremely unbalanced distribution. This weakly informed prior thus represents an \textit{optimistic} belief state in which the prior probability of any given object being in focus is artificially boosted.

\end{description}

% Our CSF model is evaluated using the density equation \eqref{eq:1} as mentioned previously. The $p(S_o^{t-1})$ term is assumed to have the distribution of either 0.33 for each of the possible cognitive statuses I, A, F (for uninformed prior probabilities) or I=0.05, A=0.1, F=0.85 (for informed prior probabilities), for the first iteration of the equation for each new object considered. We consider naming the statistical model as \textit{U-model} (with uninformed prior) and \textit{I-model} (with informed prior). $p(L_o^{t})$ is obtained from the appearance feature annotation and $p(S_o^t \mid S_o^{t-1}, L_o^{t})$ is obtained for each object in each monologue by retraining the model with all the data except the object and monologue for which it is calculated. 

% The predicted cognitive status of an object (data point) is compared with the majority vote by people about the presumed cognitive status of that particular data point. Every time there is a match we consider the model's prediction to be correct otherwise not. The number of correct predictions by the model out of all the predictions gives us the accuracy measure of the model.

%\subsection{Theoretical Models}
In addition to these two prediction vectors produced by different parameterizations of our CSF model, we also computed prediction vectors for two baseline models:

\begin{description}
\item[Finite State Machine:] First, we computed the decisions made by a rule-based FSM model, which formalized a set of heuristics from the GH coding protocol (the same heuristics previously used in the work of~\cite{williams2018reference}). In this FSM, the states correspond to cognitive statuses, and transitions are triggered based on linguistic statuses observed in incoming utterances. For example, for an FSM dedicated to some object, if that object is mentioned in a topic role, this will deterministically trigger a state transition to \textit{in focus}.
\item[Random Baseline:] Second, we computed the decisions made by a random baseline (RB) model, which predicted cognitive statuses at random. 
\end{description}

\subsection{Results}
The overall accuracy of each model (i.e., the proportion of correct entries in each model's prediction vector) is shown in Table \ref{table:accuracy}. This demonstrates that our U-model had the highest accuracy, and that our I-model and the theoretical FSM model had the same accuracy, slightly less than the U-model. The accuracy measure of the FSM model suggests that the heuristics encoded in the GH coding protocol are a good representation of the patterns that can be learned from the data we collected, given our choice of data annotations. The similarity of the CSF model's accuracy to that of the FSM similarly demonstrates that the CSF did a good job of automatically learning these patterns from our data. The slightly higher accuracy of the U-model over the I-model suggests that the uniformly distributed prior probabilities may have been more helpful than the weakly informed prior distribution. Finally, the performance advantage of all of these models over the RB model provides a good baseline measurement of success.

\begin{table}[H]
\begin{center} 
\caption{Accuracy measure of each model} 
\label{table:accuracy} 
\vskip 0.12in
\begin{tabular}{ll} 
\hline
model    &  accuracy \\
\hline
U-model       &   82.03 \\%78.12
I-model   &   81.25 \\ %78.91
FSM           &   81.25 \\
RB           &   32.81 \\
\hline
\end{tabular} 
\end{center} 
\end{table}

%Similarly, we compared the judgement vectors yielded by this process mentioned previously for each model, using a set of pairwise McNemar's Test as explained in the following section.

To validate these intuitive assessments, we formally compared our four models using six pairwise McNemar's Tests \citep{mcnemar1947note, bostanci2013evaluation}, whose results are shown in Tables \ref{table:contingency_pairs} and \ref{table:mcnemars}.

\begin{table}[H]
\begin{center}
\caption{Contingency Table entries for model pairs}
\label{table:contingency_pairs} 
\vskip 0.12in
\begin{tabular}{llllll} 
\hline
\textbf{$model_{1}$} & \textbf{$model_{2}$} & $N_{ss}$ & $N_{sf}$ & $N_{fs}$ & $N_{ff}$ \\ \hline
U-model & I-model & 104 & 1 & 0 & 23\\
U-model & FSM & 89 & 16 & 15 & 8 \\
U-model & RB & 34 & 71 & 8 & 15 \\
I-model & FSM & 89 & 15 & 15 & 9 \\
I-model & RB & 33 & 71 & 9 & 15 \\
FSM & RB & 34 & 70 & 8 & 16 \\\hline
\end{tabular}
\end{center}
\end{table}

\begin{figure}[h]
\pgfplotsset{width=9cm, compat=1.14}

\begin{tikzpicture}[
  every axis/.style={ % add these settings to all the axis environments in the tikzpicture
    ybar stacked,
    ymin=0,ymax=110,
    x tick label style={rotate=45,anchor=east},
    symbolic x coords={
      U vs FSM,
      U vs RB,
      U vs I,
      FSM vs RB,
      I vs FSM,
      I vs RB
    },
  ylabel={\# data points},
  bar width=8pt
  },
]
\begin{axis}[bar shift=-7pt, 
    hide axis,
    legend style={at={(0.5,-0.25)},
     anchor=north,legend columns=-1},
    legend entries={both correct, single correct}
]
\addplot+[ybar] plot coordinates
{(U vs FSM,89) (U vs RB,34) (U vs I,104) (FSM vs RB,34) (I vs FSM,89) (I vs RB,33)};
\addplot+[ybar] plot coordinates
{(U vs FSM,16) (U vs RB,71) (U vs I,1) (FSM vs RB,70) (I vs FSM,15) (I vs RB,71)};
\end{axis}

% and bar shift +10pt here
\begin{axis}[bar shift=1pt]
\addplot+[ybar] plot coordinates
{(U vs FSM,89) (U vs RB,34) (U vs I,104) (FSM vs RB,34) (I vs FSM,89) (I vs RB,33)};
\addplot+[ybar] plot coordinates
{(U vs FSM,15) (U vs RB,8) (U vs I,0) (FSM vs RB,8) (I vs FSM,15) (I vs RB,9)};
\end{axis}
\end{tikzpicture}
\centering
\caption{Comparison between models}
\label{fig 3}
\end{figure}
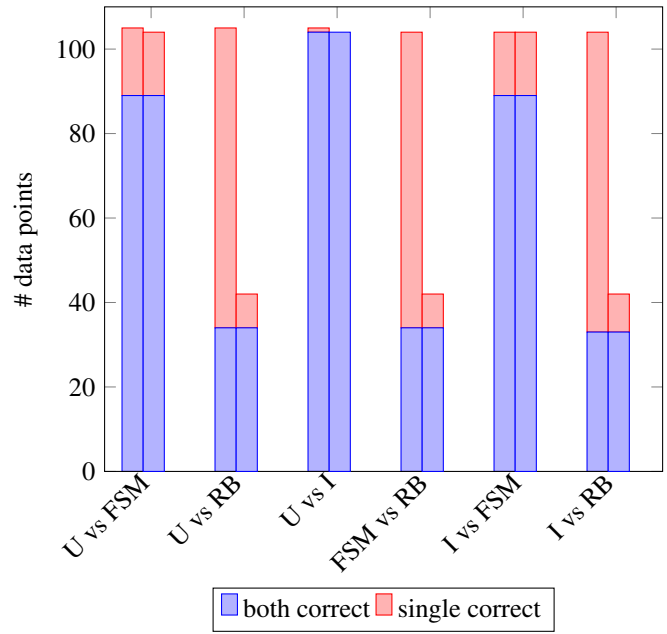

Table~\ref{table:contingency_pairs} (see also Figure \ref{fig 3}) shows the contingency table values used by McNemar's test for each pairwise comparison, where the four $N$ counts refer to the contingency table cells shown in Table~\ref{table:2x2contingency}. That table layout simply depicts a  general 2x2 contingency table \citep{liddell1976practical, clark1999performance} comparing the performance of two models A and B. Here, $N_{ff}$ and $N_{ss}$ respectively denote the number of instances where both models failed and succeeded. $N_{fs}$ and $N_{sf}$ respectively denote the instances where one model failed and the other succeeded. 

% When comparing the performance of two models as shown in Table \ref{table:mcnemars}, the $N_{fs}$, $N_{sf}$ measures are more informative than $N_{ff}$, $N_{ss}$ \cite{bostanci2013evaluation}. Thus, the contingency values calculated for all the model pairs is shown in Table \ref{table:contingency_pairs} and a visual representation of this is depicted in Figure \ref{fig 3}.

\begin{table}[H]
\begin{center} 
\caption{A 2X2 Contingency Table} 
\label{table:2x2contingency} 
\vskip 0.12in
\begin{tabular}{lll} 
\hline
    &  model A success & model A fail \\
\hline
model B success       &   $N_{ss}$ & $N_{sf}$\\
model B fail   &   $N_{fs}$ & $N_{ff}$ \\
\hline
\end{tabular} 
\end{center} 
\end{table}

The McNemar's Test statistics $\chi^2$ (with 1 degree of freedom) and p-values \citep{bostanci2013evaluation,fay2011exact,liddell1976practical} are calculated for each pair of models as shown in Table \ref{table:mcnemars}. By looking at the McNemar's Test results the following deductions can be formally made: 

\begin{enumerate}
    \item The U-model and I-model show similar performance ($\chi^2\approx$  0 and p-value = 1).
    \item The U-model and FSM also show similar performance ($\chi^2\approx$  0 and p-value = 1).
    \item The FSM and RB models show significant difference in their performance ($\chi^2$ = 47.705 and p-value = 0.0001).
    \item The CSF model and RB model differ significantly in performance regardless of model parametrization.
    \item The performance difference between the CSF model and the FSM model is not statistically significant.
\end{enumerate}

\begin{table}[H]
\begin{center}
\caption{McNemar's Test statistic ($\chi^2$) and p-values}
\label{table:mcnemars} 
\vskip 0.12in
\begin{tabular}{lll}
\hline
& $\chi^2$ & p-value\\ 
\hline
U-model, I-model & 0.000 & 1.000\\
U-model, FSM & 0.000 & 1.000\\
U-model, RB & 48.658 & $<$0.0001\\
I-model, FSM & 0.033 & 0.8551\\
I-model, RB & 46.513 & $<$0.0001\\
FSM, RB & 47.705 & $<$0.0001\\
\hline
\end{tabular}
\end{center}
\end{table}

%\section{Limitations}

\section{Conclusion and Future Work}

In this paper we present the notion of a \textit{Cognitive Status Filter}: a statistical model for estimating the cognitive status of some entity that may be referenced in conversation.  We then described a Mechanical Turk experiment used to gather ground truth data to train this model, and demonstrate how the accuracy of this model compares to a rule-based FSM model and a random baseline.

The overall accuracy of our CSF model in predicting the cognitive status of an object was slightly better than that achieved by a FSM. This simultaneously speaks in favor of the heuristics encoded in the GH coding protocol, while also demonstrating that those heuristics are learnable from data. However, there are a number of directions for future work that may significantly improve the potential performance of the statistical CSF model over the rule-based FSM model. 

First, this experiment used a relatively small corpus collected in a single task; given the fact that our model works on this small dataset one follow up step would be to collect a larger dataset from a broader set of HRI scenarios (preferably a gold-standard English corpus), as that could yield a model with better generalizability. Second, our model currently only uses linguistic status information that is already explicitly called for by the subset of the GH coding protocol used to design the FSM model. However, the CSF model could straightforwardly be extended to include additional non-linguistic cues like gaze and gesture which are critical in both human-human and human-robot communication (e.g., for establishing joint attention \citep{moore2014joint, peeters2014interplay}), which although not well described in the GH coding protocol would clearly play a role in informing notions of cognitive status. Similarly, we considered only three simple linguistic features (topic mentioned, mentioned, and not mentioned) given by the GH coding protocol, whereas more complex and varied linguistic features could improve performance. Finally, one of the theoretical advantages of the CSF model is its ability to handle uncertainty. This will be critical for integrating gaze and gesture, which are inherently ambiguous and uncertain cues.

In addition, one limitation of our experimental paradigm is that users may have been coerced into selecting an object in the scene as a candidate referent for ``it" (question Q1, i.e., as opposed to selecting nothing at all) even when they believed that no felicitous referent existed. This could be addressed in future work by modifying the question asked to participants in order to allow them to not select any present object if they did not believe them to be sufficiently likely candidates.

Finally, in future work, we intend to leverage our CSF model to implement a GH-theoretic anaphora generation model that uses an object's cognitive status when selecting a referring form during natural language generation. We further plan to integrate this model into the DIARC cognitive robotic architecture~\citep{scheutz2019overview} and demonstrate its use in realistic HRI scenarios. 

\section{Data Availability}
Our experimental data can be found at https://osf.io/qse7y/, along with our analysis scripts, experimental materials, and model outputs.

\bibliographystyle{apacite}

\setlength{\bibleftmargin}{.125in}
\setlength{\bibindent}{-\bibleftmargin}

\bibliography{CogSci_Template}

\end{document}